# Automated Building Image Extraction from 360° Panoramas for Post-Disaster Evaluation

Ali Lenjani[1], Chul Min Yeum[2*], Shirley Dyke[1,3], Ilias Bilionis[1]

*[1]School of Mechanical Engineering, Purdue University, West Lafayette, IN 47907, United States*
*[2]Department of Civil and Environmental Engineering, University of Waterloo, ON, N2L 3G1, Canada*
*[3]Lyles School of Civil Engineering, Purdue University, West Lafayette, IN 47907, United States*

**Abstract:** *After a disaster, teams of structural engineers collect vast amounts of images from damaged buildings to obtain new knowledge and extract lessons from the event. However, in many cases, the images collected are captured without sufficient spatial context. When damage is severe, it may be quite difficult to even recognize the building. Accessing images of the pre-disaster condition of those buildings is required to accurately identify the cause of the failure or the actual loss in the building. Here, to address this issue, we develop a method to automatically extract pre-event building images from 360º panorama images (panoramas). By providing a geotagged image collected near the target building as the input, panoramas close to the input image location are automatically downloaded through street view services (e.g., Google or Bing in the United States). By computing the geometric relationship between the panoramas and the target building, the most suitable projection direction for each panorama is identified to generate high-quality 2D images of the building. Region-based convolutional neural networks are exploited to recognize the building within those 2D images. Several panoramas are used so that the detected building images provide various viewpoints of the building. To demonstrate the capability of the technique, we consider residential buildings in Holiday Beach, Texas, United States which experienced significant devastation in Hurricane Harvey in 2017. Using geotagged images gathered during actual post disaster building reconnaissance missions, we verify the method by successfully extracting residential building images from Google Street View images, which were captured before the event.*

## 1 INTRODUCTION

In the aftermath of the recent severe hurricanes across Florida, the Gulf of Mexico and Puerto Rico, and the devastating clusters of earthquakes in Mexico, significant causalities and economic losses resulted from failures of the built environment. Millions of residents were forced to rebuild or repair their buildings (Smith, 2018). Understanding the nature of the risk to our building inventory is essential for making decisions that reduce risk, mitigate losses, and enhance the ability of a community to recover after a disaster.

There is strong agreement in the engineering community that engineers can and should learn much more from the consequences of each disaster than we do today (Gutmann, 2011). As part of these procedures to enable the learning process, post-event reconnaissance teams are dispatched after disasters to investigate damaged buildings and to collect perishable data (Sim et al., 2016; NCREE, 2016; Brando et al., 2017; Kijewski-Correa et al., 2018; Roueche et al., 2018). For instance, Fig. 1 shows samples of images collected from actual post-hurricane reconnaissance missions (Katrina in 2005 and Harvey in 2017). During a typical mission, images are collected from many buildings, and these images are used to document findings and observations in the field. Meaningful scenes are captured from particular viewpoints to provide visual evidence of damaged/undamaged buildings or their components. Such data allow researchers to distil important lessons that will improve the safety and reliability of our buildings, as well as the resilience of our communities (Jahanshahi et al., 2009; Gao and Mosalam, 2018; Cha et al., 2017; Cha et al., 2018; Hoskere et al., 2018; Liang, 2018; Li et al., 2018; Xue and Li, 2018; Yeum et al., 2018b; Xu et al., 2019). In the United States, the National Science Foundation recently funded a unique RAPID facility within the NHERI (Natural Hazards Engineering Research Infrastructure) network that is dedicated to supporting reconnaissance teams as they collect such field data (NSF, 2014).

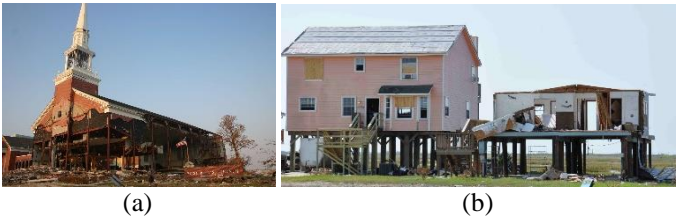

(a)                           (b)

**Figure 1.** Sample images collected during post-event building reconnaissance missions: (a) Hurricane Katrina in 2005, and (b) Hurricane Harvey in 2017 (Courtesy of Timothy P. Marshall and Thomas P. Smith, respectively).

To understand the probable causes of damage, it is essential to use the post-disaster images together with pre-disaster views of the original condition of the structure. Suppose that the target buildings are severely damaged, like those in Fig. 1. In this case, pictures of remnants of building frames and associated debris do not have significant value unless information regarding their pre-event condition is also provided. Pre-disaster building images contain abundant information regarding building-level characteristics (e.g., number of floors, architecture/structural style) and details about components (e.g., roof style, windows opening, column locations). This information is critical for understanding the root causes of damage during the event (Suppasri et al., 2013).

Recent advances in sensors, sensing platforms, and data storage enable automated solutions for structural engineering problems (Rafiei and Adeli, 2017; Choi et al., 2018; Gao and Mosalam, 2018; Hoskere et al., 2018; Yeum et al., 2015, 2018a, 2018c). One of the best resources available to observe buildings in their pre-disaster condition is street view images (Anguelov et al., 2010). Through such services, a sequence of 360-degree panorama images (hereafter, panoramas) captured along nearby streets enables all-around views at each location. Many internet-based map service providers offer a street view service, such as Google or Bing in the United States or Tencent in China, and these service providers store a sufficiently broad temporal and geographic range of panoramas. Using the GPS coordinates of a particular building, several external views of that building can be extracted. Thus, facilitating the realistic use of such data is expected to yield much more granular information regarding the pre-event state of the buildings in a region.

In this study, we develop an automated technique to detect and extract curated pre-event building images from typical street view panoramas. This technique is intended to support research investigating the impact of disasters on a building inventory. The extracted images capture the external appearance of a building from several viewpoints. As a preliminary step, a classifier is first trained to detect buildings within images using a large ground-truth building image set. The region-based convolutional neural network algorithm is exploited to design a robust building classifier (Ren et al., 2017). Ideally, in the real-world application of this technique, a user can simply provide, as the input, a geotagged image (or similarly, a GPS coordinate) recorded near the target building, and the rest of the process is fully automated. The physical location of the target building is estimated using the geometric relationship between the panoramas and the building. Then, the optimal projection plane for each of the panoramas is determined to produce a high resolution, undistorted 2D building image from the corresponding panoramas. The region (position) of the building on each 2D image is determined using the trained classifier, and an image of the building is extracted from each 2D image. The output of the technique is a set of several undistorted images of the building, taken from all available viewpoints. Generating multiple image reduces the possibility of an obstruction (e.g. trees, cars or fences) in a particular viewpoint and, thus, enables robust visual assessment. Also, by using this procedure with data from multiple street view services, a user can collect street view images obtained over many past years to observe the target building over time. The performance of the technique developed here is demonstrated and validated using residential buildings affected by Hurricane Harvey (in 2017) in Holiday Beach, TX. Real-world post-event reconnaissance images, collected by engineers in the field, are used as the input for this validation, and pre-disaster views of the buildings in those reconnaissance images are automatically extracted from Google Street View (Shet, 2014).

The major contribution of this study is to provide a practical and feasible solution to a problem that is grounded in the needs of engineers who seek to learn from disasters. With the capability to analyze post-disaster building scenarios by readily accessing pre-disaster images and post-disaster data, engineers are equipped with the tools to rapidly develop a greater understanding of the performance of our infrastructure. The technique developed automatically provides high-resolution undistorted multiple view pre-event images of each building. The key technical contributions are in automatically removing the inherent distortion of the 2D projection of panorama images and in rapidly extracting a set of images of each post-disaster target building with multiple viewpoints. Additionally, a side benefit of this technique is the ability to exploit the vast amounts of legacy visual data that exist from past disasters. With such visual databases being collected and established, and the cost of image acquisition and data storage decreasing, this technique offers just one example of how to automate the reuse of existing data through the novel application of state-of-art computer vision algorithms to solve real-world problems.

The paper is structured as follows. First, the challenges in using street view images for our application are introduced in Section 2. Section 3 provides a brief introduction of the convolutional neural network-based object detection methodology, called region-based convolutional neural network (Ren et al., 2015). The technical details of the technique are explained in Section 3, and in Section 4, it is demonstrated using Google Street View images gathered from Holiday Beach in Texas, United States, which was heavily affected by Hurricane Harvey in 2017. The summary and conclusions of this study are presented in Section 5.

## 2 Problem Statement

The 360-degree panorama image is an image that is able to capture all around (spherical) scenes from a given location. The merit of panoramas is that, after the data are collected, one may quickly navigate to scenes of interest in any direction from the given location. This advantage does not apply to 2D images. With 2D images, the direction in which the data are being collected must be determined at the data collection stage. Thus, panoramas are quite useful when for reconnaissance because when they are collected from multiple locations, several different external views of a given building are automatically recorded (e.g., front or sides). In the panorama viewers typically implemented by street view service providers, 2D rectilinear images are rendered from the panoramas in real-time based on the selected viewing direction (represented by a pitch and yaw) and zoom level (Google Developers, 2018a). When rendered in this way, the rectilinear images are just the 2D images that would typically be obtained with an ordinary (non-fisheye) camera. These images represent scenes in the world as people actually see and perceive them. For instance, a straight line in the 3D world is represented as a straight line in the corresponding rectilinear image (Kweon, 2010). Such viewers also frequently overlay directional arrows to enable a user to move to nearby panoramas.

Although such viewers provide accessible, easy-to-use and easy-to-view panoramas, a great deal of manual effort is now required to retrieve data for a particular building. First, the user must read the GPS coordinate from a geotagged image and enters the GPS coordinates into the viewer. Then, by panning and zooming, the user finds the target building and determines the best perspective for a clear view the front of the building. The user captures the rectilinear image shown on the screen and crops the image to extract and save the building region to document this view. Next, to observe the side of the building (or any other angle), the user must click on the directional arrows to move to nearby panoramas and repeat the extraction procedure. This entire process is repeated for each of the panoramas (locations) yielding several images of the building from various viewpoints. This manual process is time-consuming and inefficient. Gathering multiple pre-disaster images for a large number of buildings in a subdivision or city would take a great deal of time for a human.

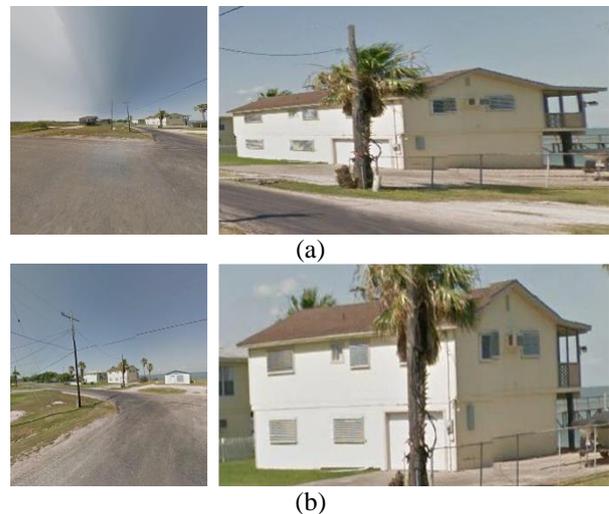

(a)

(b)

**Figure 2**. Rectilinear images created from an (a) incorrect and (b) correct viewing angle (direction of projection): These images are generated from the same panorama (Google Street View, 2018).

To automate this process, we incorporate two key capabilities into the technique developed. First, the position of the building on each image is identified. Herein, to avoid confusion, we use the term 'position' when we mean the location of the building on the 2D images. The term 'location' is used only for its geolocation in the world (3D). We exploit a recently developed deep convolutional neural network algorithm, which has led to breakthroughs in object recognition (Ren et al., 2015). Using a large number of labeled building images, a robust building classifier is trained to accurately identify the presence of the building and detect its position in the pictures. The details of this algorithm are explained in Section 3. Next, the optimal viewing direction (direction of projection) is determined. The direction of projection is defined as the viewing direction that generates a rectilinear image from the panorama to yield an undistorted view in the selected direction. However, inaccurate selection of the direction of projection may result in significant distortion of the target building. For example, Fig. 2 demonstrates the effect of projection direction on rectilinear images. Both sets of images are generated from the same panorama. When the target building is located further from the center of a rectilinear image, its quality is degraded due to a large distortion. Thus, the most suitable direction of projection of each panorama should be determined to generate the 2D rectilinear image with the best available quality. The implementation of this capability is presented in Section 3.

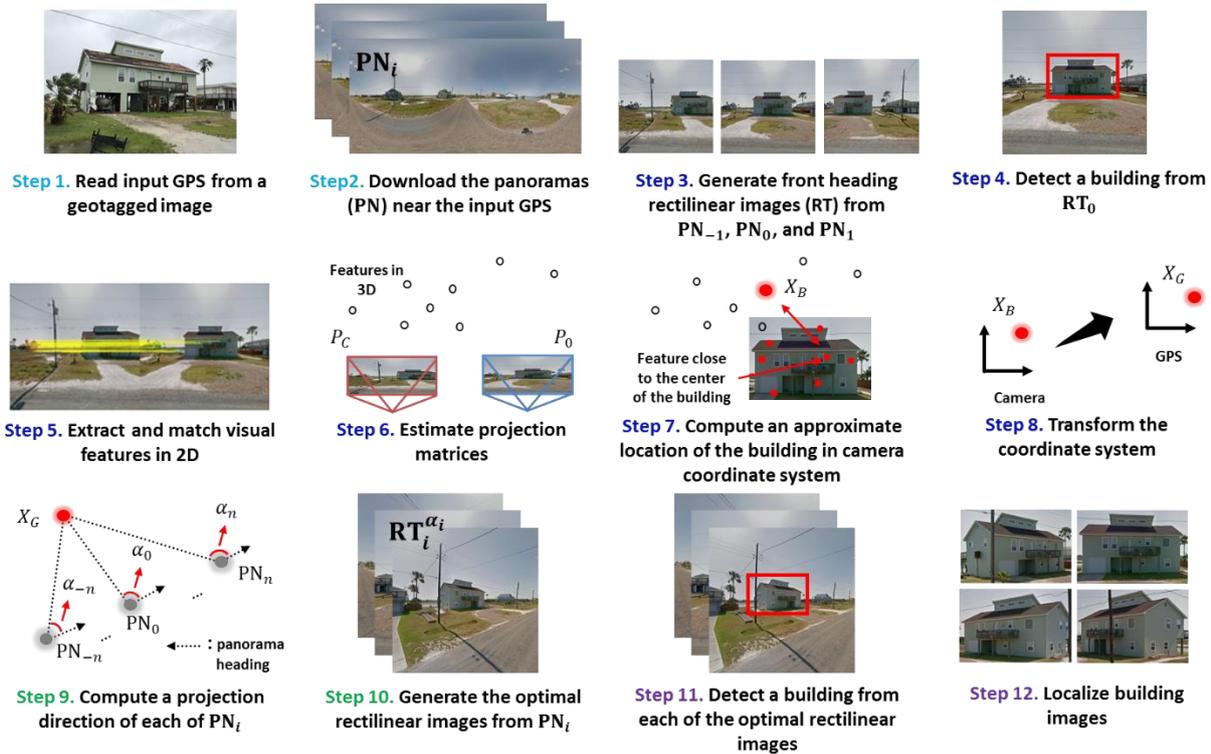

**Figure 3.** Overview of the technique developed

## 3 Methodology

The technique developed herein is explained in Fig. 3. The details are as follows. Users provide a geotagged image of the target building (which includes the GPS coordinate near that target building) and the input, and several images of the corresponding building are automatically extracted from street view images and the outputs. The resulting images contain views of the building acquired from various viewpoints and with negligible distortion. This technique will enable the user to readily examine images of the entire front and sides of the building in its pre-event condition.

The overall technique is divided into four main processes: Steps 1 and 2 (captioned in light blue) download all panoramas acquired near the target building from the street view service. Steps 3 to 8 (captioned in dark blue) approximate the building location in the GPS coordinate system using the geometric relationship between the building and the panoramas. Rectilinear images are generated from a couple of the closest panoramas, and the trained building detector is applied to find the building position on the rectilinear images. Then, the location of the building is identified in each of the camera coordinate systems, and transformed into the GPS coordinate system. Note that this intermediate step is not yet intended to extract high-quality building images because the rectilinear images are generated without considering the optimal direction of rectilinear projection. Steps 9 to 10 (captioned in green) are generate the rectilinear image from each of the panoramas by considering its optimal direction of projection. Lastly, Steps 11 to 12 (captioned in purple) detect the target building in each of the optimal rectilinear images using the same building detector and extract the building images for the final use. Again, once a geotagged image is provided in Step 1, the rest of the process is fully automated. The details of each step are provided below:

In Step 1, the GPS coordinate near the target building (hereafter, the input GPS coordinate) is obtained from the EXIF metadata of the input geotagged image(s) (Google Maps, 2018). Note that this image is only needed for providing the GPS information near the target building and thus, its quality or visual contents do not affect this technique. Alternatively, a user may manually provide approximate GPS information for the target building.

Step 2 is to download a sequence of panoramas acquired near the input GPS coordinate, denoted as $PN_i$ ($i = -n, ..., +n$), from the Street View API. Herein, $i = 0$ represents the index for the panorama that is closest to the building, and $i = -n$ and $i = +n$ represent the indices for the far left and far right panoramas in the selected set, respectively. The sequence to be downloaded begins with $PN_{-n}$ where the nearest available panorama is set to $PN_0$. For example, we assign that $PN_1$ becomes the panorama closest to $PN_0$ in one direction along the route where the panoramas are collected. Here, $n$ is the maximum allowable number of panoramas to be used in one direction. The user may initially select this number. Panoramas that are captured far from the input GPS coordinate are not so useful because the building will be too small on the image. Thus, a reasonable quantity should be set (e.g., 3 or 5). Note that, depending on the building location, the number of available panoramas may not be $2n + 1$. For instance, a building located close to a dead-end or a cul-de-sac will have fewer panoramas available in one direction. Each downloaded panorama contains panorama heading information in the form of an angle with respect to North (see Fig. 4) which is necessary for finding the optimal direction of projection. In Google Street View, since this panorama heading is set to the driving direction of the data collection vehicle, the panorama heading of a given panorama is closely aligned with the direction of the street.

In Step 3, front projection rectilinear images, denoted $RT_i^\alpha$, are generated from $PN_i$ ($i = -1, 0, 1, \alpha = 90°$ or $270°$). As mentioned in Section 2, a rectilinear projection is one type of projection to reproduce 3D scenes using 2D images, and results in the most natural way for the viewer (Ruder et al., 2018). The scenes (images) available in typical street image viewers are rectilinear images that have been projected to the direction that the user selects. This direction corresponds to the direction of projection in Fig. 4 (Google Street View, 2018). Rectilinear images are produced by mapping the panorama scene (here, panorama scene shown along the curve from A to B) to the rectilinear image plane $\overline{AB}$. Due to the available quality of street view imagery, which reduces the chance of imperfection in panoramas, the projection error is negligible. $\overline{AB}$ is determined by defining a projection angle ($\alpha$) and a field of view ($\theta$). Thus, an infinite number of rectilinear images can be generated from a single panorama by using different projection directions. $PN$ in Fig. 3, and $\overline{AB}$ in Fig. 4 are represented by pixels. The field of view $\theta$ should generally be less than $120°$ because a large field of view produces large distortion in the contents of the scene at the edges of the image. In this step, only those panoramas immediately adjacent to $PN_0$ ($PN_{-1}$ and $PN_1$) are used for computing the building location in the GPS coordinate system. We only consider $PN_i$ ($i = -1, 0, 1$) because the corresponding $RT_i^\alpha$ are most likely to include the target building ($i = -1, 0, 1, \alpha = 90°$ or $270°$). Since the true GPS coordinate for the building is unknown, the precise direction of projection for each $RT_i^\alpha$ cannot be determined. Here, we reasonably assume that the building is located along the route of the street where the panoramas are captured. Thus, the direction of projection is approximately set to either $90°$ or $270°$, making it perpendicular to the direction of the panorama heading (the direction of the street). Depending on the location of the input GPS coordinate with respect to the panorama location and heading, the direction of projection is selected to be either $90°$ or $270°$, here denoted as $p$. Note that this is not the projection direction used for extracting the final building images. Thus, images extracted using this projection direction are referred to here as *front heading rectilinear images*, $RT_i^p$.

Herein, the actual projection direction for generating rectilinear images is represented by two angles, azimuth and pitch. However, we only consider the azimuth angle in this study, which is illustrated in Fig. 4. Thus, $\theta$ and $\alpha$ are the viewing and projection "azimuth" angles, respectively. Accordingly, $\overline{AB}$ becomes the projection plane in the horizontal direction. For the vertical direction, the pitch angle of the projection is set to zero, and $\theta$ for the vertical direction is selected to be identical to the value for the horizontal. In low-rise buildings, since the optimal projection pitch angle is relatively small compared to the azimuth angle, the pitch angle does not need to be controlled, and when it is set to zero, distortion in the rectilinear linear image is insignificant. However, in the case of high-rise buildings, the pitch angles should also be controlled so that the view of the building in a vertical direction can be fully included in a corresponding rectilinear image.

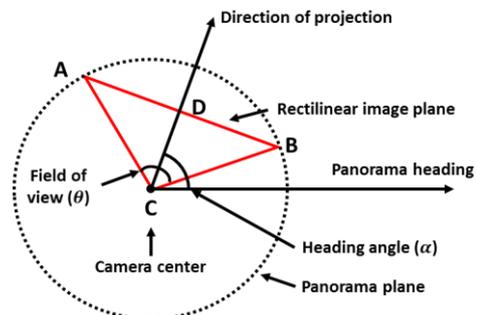

**Figure 4.** Geometry of a panorama and rectilinear image

In Step 4, the position of the target building in $RT_0^p$ is detected using a trained building (object) classifier. This building classifier is trained in advance using a large volume of ground-truth building images. The details of the object detection procedure are explained in Sections 3 and 5.1. The same trained classifier is also used later in the technique in Step 11. The ground-truth building images used for training the classifier must include images of buildings that have a similar appearance as the target building. For instance, if the building images used for training only contain wooden buildings, the classifier may not be sufficiently accurate when classifying images of masonry or concrete buildings (Yeum et al., 2018b). After applying the classifier to $RT_0^p$, we determine a tight bounding box for the building in that image. If more than one building is detected in $RT_0^p$, the bounding box that is closest to the center of $RT_0^p$ is selected, denoted as $bbox_0$ (the subscription indicates the order, or index, of the image).

Step 5 is to extract and match the visual features between $RT_i^p$ ($i = -1, 0, 1$). The features extracted from $RT_0^p$ must be matched with the corresponding features from $RT_{-1}^p$ and $RT_1^p$. Conventional visual features and descriptors (e.g., SIFT or SURF) can be used for this process (Lowe, 2004). These visual features and their descriptors represent unique key points across the images and are used for computing the geometric relationship (essential matrix in this study) between images. Then, a single rectilinear image is selected from either $RT_{-1}^p$ or $RT_1^p$, which is chosen as the one with the larger number of matched features, denoted as $RT_c^p$. This process is intended to improve the accuracy of building location estimation by using more information. For example, the building on $RT_0^p$ is often shifted slightly because the input GPS coordinate may not be recorded at the location of the panorama closest to the house. In such a case, either $RT_{-1}^p$ or $RT_1^p$ may be far from the panorama closest to the building, and the other panorama is unlikely to include the same building, causing a failure in estimating the building location. Unless the building is far from the street along which the panoramas were captured, the building often occupies a large portion of each image, and thus many features generated from the building regions will be matched. One of these features that is close to the center of $bbox_0$ will be used for computing the building location in Step 7.

In Step 6, the projection matrices of $RT_0^p$ and $RT_c^p$ are computed. The projection matrix is a 3×4 matrix which represents the mapping from 3D points in the world to 2D points in an image. In this study, the projection matrices of $RT_0^p$ and $RT_c^p$ are computed using an essential matrix between two images, and the corresponding projection matrices are denoted as $P_0$ and $P_c$ in (see Fig. 3), respectively. The essential matrix is a special case of a fundamental matrix for which the intrinsic (calibration) matrix for both cameras is known (Hartley and Zisserman, 2003). Using the panorama's geometry in Fig. 3, the intrinsic matrix having principal points and a focal length is obtained, and it is identical for both $RT_0^p$ and $RT_c^p$. Since the essential matrix has only five degrees of freedom to be estimated from $RT_0^p$ and $RT_c^p$, fewer degrees of freedom than the fundamental matrix, more accurate projection matrices can be computed. Based on the pairs of visual feature matches in Step 5, the essential matrix is estimated using the five-point algorithm combined with RANSAC (RAndom SAmple Consensus) (Fischler and Bolles, 1981; Nister, 2004). For the mathematical representation of this process, the intrinsic matrix ($K$) is represented as:

$$K = \begin{bmatrix} f_x & 0 & p_x \\ 0 & f_y & p_y \\ 0 & 0 & 1 \end{bmatrix} = \begin{bmatrix} f & 0 & p \\ 0 & f & p \\ 0 & 0 & 1 \end{bmatrix}, \tag{1}$$

where $f$ and $p$ are the focal length and the coordinates of the principal point, and the subscripts $x$ and $y$ indicate the width and height directions, respectively. Since we consider the same viewing angle in the horizontal and vertical directions, so $f_x$ and $f_y$ (and $p_x$ and $p_y$) are the same regardless of the direction. In Fig. 4, $\overline{CD}$ and $\overline{AD} (= \overline{BD})$ are $f$ and $p$, respectively. Thus, $f$ becomes:

$$f = p \cdot \arctan(\theta / 2), \tag{2}$$

where $p$ is half of the size of the rectilinear image in pixels, meaning that the principal point is the center of the rectilinear images. The pairs of matched feature points are transformed by multiplying the inverse of $K$ by their points coordinates (in the homogeneous coordinate system). Then, those points are expressed in a normalized coordinate system, denoted as $x_0$ in $RT_0^p$, and $x_c$ in $RT_c^p$ (Hartley and Zisserman, 2003). The essential matrix ($E$) satisfies the following relationship:

$$x_c^T E x_0 = 0. \tag{3}$$

The five-point algorithm based on this relationship is utilized as a hypothesis-generator (model) for RANSAC to count the number of inliers and outliers, and the inliers are used for estimating $E$ (Hartley and Zisserman, 2003; Li and Hartley, 2006). The estimated $E$ enables us to extract $P_c$ if $P_0$ is assumed to be a canonical projection matrix ($P_0 = [I \quad 0]$ where $I$ is a $3 \times 3$ identity matrix) (Hartley and Zisserman, 2003). In this case, the origin of the camera coordinate system is the camera center (focal point) of $RT_0^p$ (generated from $PN_0$).

In Step 7, the approximate 3D locations of the building are identified in the camera coordinate system. When the direction of projection is generally aimed toward the center of the building façade, distortion on the rectilinear images can be minimized. Given the projection matrices $P_0$ and $P_c$, we can compute the 3D points in the camera coordinate system that correspond to a pair of matched image points (point correspondence) using a linear triangulation algorithm (Hartley and Zisserman, 2003). From Steps 4 and 5, we have a set of matched image points between $RT_0^p$ and $RT_c^p$. However, the visual features and/or the matched sets are randomly distributed over the building region on those images, and thus it is not guaranteed that we will obtain the exact center of the building façade. Moreover, depending on the locations of $PN_0$ and $PN_c$ with respect to the building and the width of the building, they may include a portion of the side of the building, meaning that the center of the bounding box may not be the center of the building façade. Thus, we reasonably define the 3D building location (in the camera coordinate system) as the 3D point generated from the matching feature closest to the $bbox_0$. Once we find the matching feature nearest to this point, its corresponding location in 3D space is identified in the GPS coordinate system. The estimated building location is denoted as $X_B$.

Step 8 is to perform a 2D similarity transformation to define $X_B$ in the GPS coordinate system, denoted as $X_G$. In Step 2, each of the panoramas downloaded from street view services has its own accurate GPS information (Google Maps, 2018) and the 3D locations of $PN_0$, $PN_c$, and $X_B$ are computed in the camera coordinate system using Steps 3~7. However, these two control points (the same points in two different coordinate systems) are not entirely sufficient to compute a 2D similarity transformation between the two coordinate systems (Horn, 1987). Since the positions of the camera when it is acquiring the data (panoramas) are almost co-linear (because the street view vehicle is driving along a street), adding one additional control point by considering another (3rd) camera location does not add a new equation to obtain a unique transformation. Alternatively, we choose to eliminate one of the dimensions in both coordinate systems. Because the panoramas are captured from a street view vehicle, the camera acquiring the panoramas moves along a path that is almost entirely in a single plane with a consistent height (negligible street slope exists between two panorama locations) in the camera coordinate system. Accordingly, we can assume that the variation in the camera's altitude in the GPS coordinate system is minimal and can be ignored. With two pairs of camera locations in both the camera coordinate system (considering only the image width and depth directions) and the GPS coordinate system (considering only latitude and longitude), we have a sufficient number of equations to perform the 2D similarity transformation using the two translational, two rotational, and one scaling parameters that are available. Here, the camera location in $PN_0$ is $[0 \quad 0 \quad 0]$ and $PN_c$ becomes $-M^{-1}p_4$ where $M$ is the left $3 \times 3$ submatrix of $P_c$ and $p_4$ is the last column of $P_c$ (Hartley and Zisserman, 2003). GPS coordinates are typically recorded in geodetic coordinates (represented as longitude and latitude). Note that this system is not a Euclidean space, as the one used in the camera coordinate systems. The geodesic datum should thus be represented by the equivalent values in Euclidean space (Drake, 2002). Since the distance between the two panoramas is relatively small compared to the radius of the Earth, we can use the flat Earth assumption in which the values are transformed into an Earth-north-up (ENU) coordinate system (Pandey et al., 2011). The ENU coordinates are formed from a plane tangent to a fixed point on the Earth's surface. Coordinates of the points are found by computing translational movements on the tangential plane to East (X) and North (Y) from the fixed point. Then, finally, we can identify the 2D similarity transformation matrix using these two pairs of control points defined in two different Euclidian coordinate systems. The transformation matrix is then applied to $X_B$ to obtain $X_G$.

In Step 9, the correct projection directions for $PN_i$ ($i = -n, ..., +n$) are computed to generate the optimal rectilinear images. $PN_i$ have the panorama heading with respect to North, and their locations are defined in the ENU coordinate system obtained in Step 8. Thus, as shown in Step 9 in Fig. 3, the correct projection directions ($\alpha_i$) can be computed.

In Step 10, $RT_i^{\alpha_i}$ are generated from $PN_i$ using $\alpha_i$ computed in Step 9 ($i = -n, ..., +n$), and these are the optimal rectilinear images. This step repeats the same process in Step 3 for all panoramas, although at this point the process is performed using the correct projection angles. Since the projection direction is aimed toward the target building, the target building is now at the center of each $RT_i^{\alpha_i}$.

In Step 11, the position of the target building in each $RT_i^{\alpha_i}$ is detected using the trained building classifier, which is identical to the classifier used in Step 4. The only difference compared to Step 4 is that here we detect the building within the optimal rectilinear images generated from all panoramas, which contain undistorted views of the target building.

Finally, in Step 12, we obtain the set of highly localized target building images captured from various viewpoints. The detected target building images having various viewpoints are cropped from $RT_i^{\alpha_i}$, which are captured from different locations.

In our study, the visual recognition (Papageorgiou et al., 1998; Viola and Jones, 2004; Dalal and Triggs, 2005) of the building is used both for estimating its 3D location by computing the geometric relationship between the panoramas (Step 4 in Fig. 3) and for detecting and cropping its region on each rectilinear image (Step 11 in Fig. 3). Accurate detection and positioning of each building on the images are critical for achieving the successful extraction of pre-event building images from the panoramas. Recently several CNN based high performance object detection algorithms have been presented (Girshick et al., 2015; Ren et al., 2015; Redmon and Farhadi, 2017). We incorporate a state-of-art object detection method, called faster region-based convolutional neural network (Faster-RCNN) into our technique (Ren et al. 2015; 2017). Faster-RCNN is an evolved version of region-based convolution neural network in terms of speed and accuracy (Harzallah et al., 2009; Hinton and Salakhutdinov, 2009; Uijlings et al., 2013; Girshick et al., 2014; Chen et al., 2015; Girshick, 2015; LeCun et al., 2015; Krizhevsky et al., 2017). Recently, an enhanced version of Fast R-CNN, in terms of speed and accuracy. Many other architectures have been introduced to reduce the training and testing speed as well as the accuracy, but there is always a trade-off between the accuracy and computational efficiency (Liu et al.,2016; Redmon and Farhadi, 2017). In this study, we implement the original Faster R-CNN to detect buildings on the images.

To close this section, we comment on some assumptions used in the technique that one must remember for successful implementation. First, of course, the panoramas used must contain the target building. Updates on the panoramas in the street view are infrequent in remote areas, and this highly depends on the location (i.e., urban or rural regions), although data collection is generally increasing in frequency. Recently constructed buildings may not be captured in the panoramas, and renovated buildings may not always have up-to-date images. One possible solution is to explore various street view services as their data collection periods and frequencies are different. Second, the geotagged image used as the input to the technique should be collected from a location near the target building. The GPS coordinates stored in the metadata should be close to the target building. Recall that the actual visual contents of this input image is not actually used in the technique, and its quality is not relevant. However, it is recommended that each post-disaster geo-tagged image include only one building to prevent confusions while comparing the images with pre-disaster images. For example, if the input image is acquired from a location that is closer to a different building, images of the wrong building may be generated. Third, the technique relies highly on the availability and accuracy of information in the street view service used. Here the panoramas and related metadata are directly obtained from street view services. If such information has limited availability or it is not accurate, the results will be incorrect. For example, incomplete or erroneous GPS coordinates for the panoramas would yield the wrong a building location, followed by erroneous projection direction estimation. However, GPS data is generally accurate enough for this purpose, and we have not observed any errors in the data to date. Although we successfully demonstrate the technique using Google Street View, we have not tested it using panoramas available through the other street view services. Fourth, the panoramas acquired must have sufficient spatial coverage (roughly not more than 10 meters). If the distance between adjacent panoramas is too far to contain the same target building, the technique will fail to correctly estimate the location of the building in Steps 3 to 6. Moreover, such sparse panoramas hinder the goal to obtain several high-quality building images. Fifth, there should be a reasonable distance between the panoramas and the building, enough to ensure that the panorama does exclusively contain the building in a rectilinear image. Since most of the panoramas are captured along the street, their distances from the building are often sufficiently far from the target building (e.g., more than the width of a single lane). However, when the distance between the buildings and the panorama location and/or their height or width are large, a greater FOV is needed to capture the entire view of the building. This limitation would cause a large distortion in the rectilinear images.

## 4 Experimental Validation
### 4.1 Description of the test site

To demonstrate the performance of the technique, we use residential buildings in Holiday Beach in Rockport, Texas as our case study. Hurricane Harvey in 2017 is the second-most costly hurricane in U.S. history, inflicting $125 billion in damage, primary due to catastrophic rainfall-triggered flooding in the Houston metropolitan area (Smith, 2018). Harvey made landfall as a Category 4 Hurricane in southern Texas, and Rockport was directly in the path of Harvey, causing tremendous wind and storm surge damage (Metz, 2017). Holiday Beach, shown in Fig. 5(a), is a residential community and most of its residential buildings (more than 80%) (hereafter, houses) in the region were substantially damaged (Villafranca, 2017).

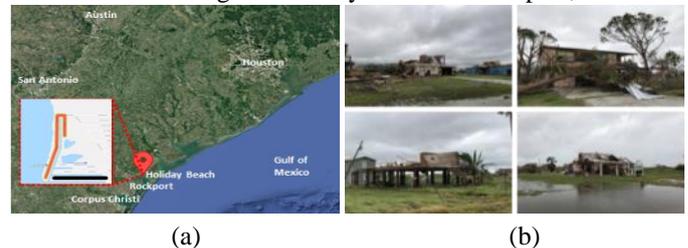

(a)                                    (b)

**Figure 5.** Test site for experimental validation, Holiday Beach: (a) Geolocation of Holiday Beach on a map and a selected path used for constructing house image database and (b) Samples of damaged residential buildings in Holiday Beach after Hurricane Harvey in 2017 (Metz, 2017).

Several reconnaissance teams were dispatched to these regions in the weeks and months after the event to evaluate structures, characterize the event, and collect data to learn from this event (FEMA, 2018). Several such teams published their data through designsafe-ci.org and weather.gov (Metz, 2017; Stark and Wooten, 2018). Geotagged images collected from Holiday Beach are also available, the selection of sample images shown in Fig. 5(b) highlights the need for observing their pre-disaster condition. It is evident that when presented with such photos of severely damaged houses, little information is available to identify the vulnerabilities that may have existed. Here, these post-disaster reconnaissance images are the input to the technique and are used solely for providing GPS coordinates to automate the process of extracting useful photos of the target houses.

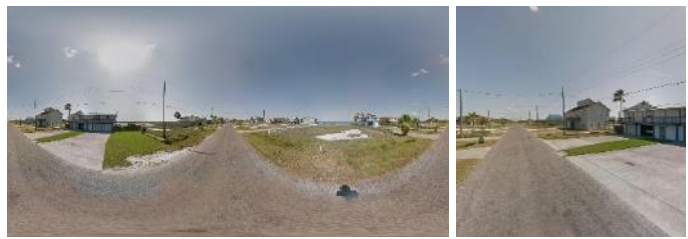

(a)                  (b)

**Figure 6.** Sample panorama in (a) and the corresponding optimal rectilinear image in (b) for a target house

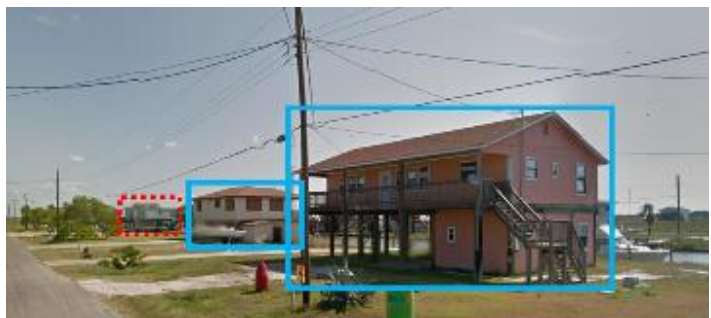 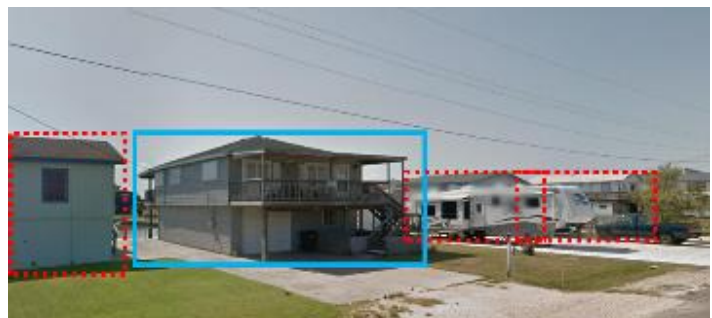

**Figure 7**. Sample building images used for training a residential building (house) classifier.

## 4.2 Construction of residential building image database

A large volume of ground-truth house images is prepared to train a residential building classifier using the algorithm discussed in Section 3. The house images used for this training process should be similar in appearance (architecture style) to those expected for actual testing and implementation. For instance, our goal here is to detect wooden residential buildings for single family residence. Wood is a standard construction material for houses across the Southern part of the United States. Training a classifier using images of high-rise apartments or concrete buildings would not improve its performance unless those are expected to be present in the images collected from the testing and implementation regions. For simplicity, we prepare our ground-truth training images using representative scenes from Holiday Beach to ensure the buildings have similar styles and appearances. The distribution of the buildings used in the ground-truth dataset is quite similar to the appearance of houses across the other coastal areas of the United States. To prepare the training images for the classifier, we exploit Steps 1 to 10 from the technique developed to generate a large volume of optimal rectilinear images. Then, we manually label that large volume of images to construct our ground-truth residential building database. Note that the resulting residential building classifier is generally applicable, and is trained in advance for use across many events and regions. The database can also be expanded over time to encompass other building types, materials, architecture and styles.

First, panoramas along the selected route on the waterfront in Fig. 5a (marked a red line) are downloaded from Google Street View. Google Street View does not provide a service to directly download high-resolution panoramas, but does have an API to download (show) a portion of the panorama when the user specifies a horizontal and vertical position and a size (Google Developers, 2018a; Google Developers, 2018b; Google Developers, 2018c). The tool that we developed automatically downloads the tiles of each panorama and stitches them together to generate high-resolution panoramas. The resolution of each panorama is $13,312 \times 6,656$ pixels. Each panorama is constructed by stitching 338 tiles having a resolution of $512 \times 512$ pixels. A total of 128 panoramas are available along these routes. A sample of a high-resolution panorama is shown in Fig. 6(a). Second, we manually select the GPS locations of many of the houses along the chosen route in Fig. 5(a). A footprint (outline) of each house can be viewed in Google Maps, and thus, its GPS location can be easily obtained. 100 houses are considered along the route having similar house styles (e.g., number of floors, roof style, elevated house foundation). Note that for constructing the ground-truth dataset, this manual process is necessary and replaces the geotagged images that would be used in the implementation of the technique following Steps 3 to 8.

Third, optimal rectilinear images are generated from the downloaded panoramas. Since here we know the GPS coordinates of each of the downloaded panoramas as well as the houses, the optimal projection angles can be directly computed. Here, $n$ is set to be five, which is the maximum allowable number of images on one side of the closest panorama location. Thus, for each house, a maximum of 11 rectilinear images are constructed from the corresponding panoramas (houses at a dead-end along the route would have fewer than 11 panoramas), and each house is positioned at the center of each of the rectilinear images. $\theta$ and $\overline{AB}$ are $120°$ and 2,048 pixels, respectively. As a result, a total of 1,056 rectilinear images having a resolution of 2,048 × 2,048 pixels are generated for the database. Figure. 6(b) shows the optimal rectilinear image generated from the sample panorama in Fig. 6(a).

Finally, the houses on the rectilinear images are manually labeled. We used a Python-based open source labeling tool to label a tight bounding box around each house (Tzutalin, 2015). Only houses that satisfy the following two conditions are labelled: (1) less than 70% of the front and side views of the house are obstructed by foreground objects, and (2) the height or width of the bounding box encompassing each house is larger than 200 pixels, which is around a tenth of the rectilinear image size in pixels. Samples of labeled houses are shown in Fig. 7. Here, only bounding boxes with solid blue lines are labeled as a house, while the other houses in these images are not included as ground-truth data. Those houses are marked with a red dotted line purely for purposes of illustrating non-labeled house data. The left image in Fig. 7, the red dotted bounding box around the

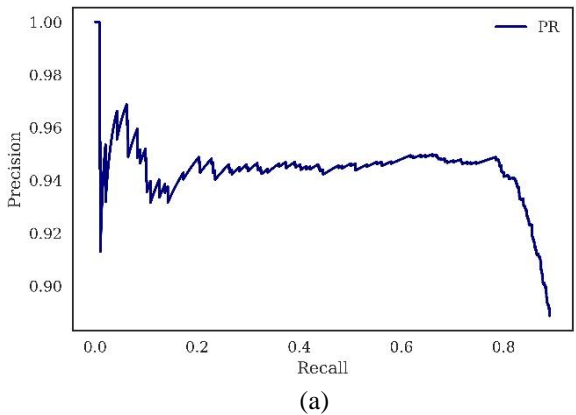
(a)

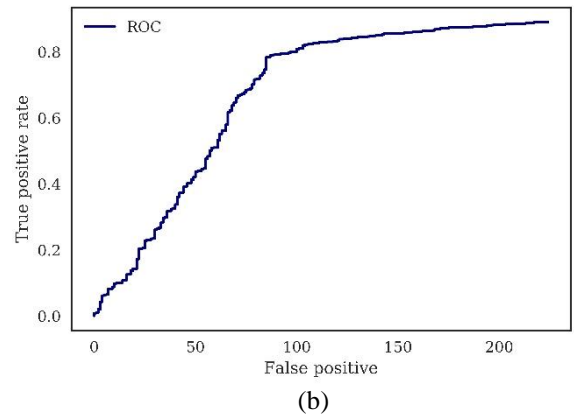
(b)

**Figure 8.** The performance of the residential building detector for the testing set: (a) PR curve and (b) ROC curve.

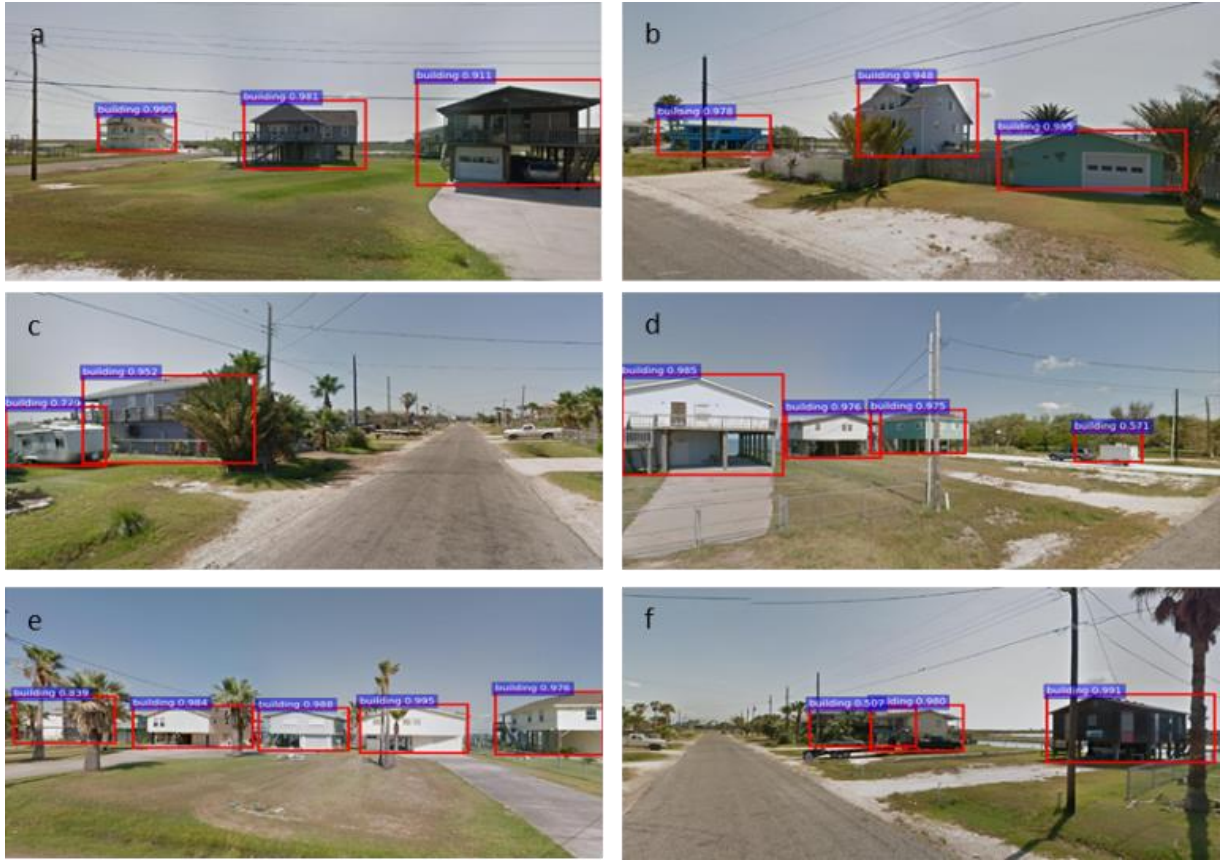

**Figure 9.** Samples of detected residential buildings using Faster R-CNN algorithm.

house on the left has height less than 200 pixels, and thus it does not contain sufficient information for our purpose. Similarly, in the right image in Fig. 7(b), all non-labeled houses are obstructed by the foreground objects (e.g., vehicle, tree, or other house) or are only partially visible. A total of 3,500 ground-truth residential buildings are labeled from the rectilinear images, and this database is utilized for training the classifier.

### 4.3 Training and testing residential building detector

An open-source library of Faster R-CNN deployed using Python is used for training and testing the residential building detector (Chen and Gupta, 2017). A single NVIDIA Tesla K80 is used for this computation. A total number of 3,500 residential houses have been labeled from 1,050 images. This set includes different views of 100 houses in Holiday Beach. Sets containing 60%, 10%, and 30% of the images of target houses are randomly chosen as training, validation, and testing sets, respectively. The Residual Network Model (called ResNet_101) is selected for learning robust features and increasing its efficiency (He et al., 2016). We manually chose the hyper-parameters for Faster R- CNN based on trial-and-error. Then, the ConvNets and Fully- connected layers are initialized by zero-mean Gaussian with standard deviations of

0.01 and 0.001, respectively. Hyperparameters including learning rate, momentum and weight decay for training R-CNN and RPN networks are set to 0.001, 0.9, and 0.0005, respectively. The learning rate is defined as the amount the weights are adjusted with respect to the loss gradient. The decay weight is set to avoid overfitting by decaying the weight proportionally with its size. We set the momentum to 0.9 to make the gradient decent achieve a faster convergence. We used three different aspect ratios (0.5, 1, and 2) and anchor sizes (128, 256, and 512 pixels). In the test stage, in cases in which a set of the detected proposals that overlap each other with greater than a 0.3 intersection over union (IoU), an additional process is needed to obtain a precise bounding box. In each set, the proposal having the highest confidence score (probability) remains, and the others are disregarded. This process is called non-maximum suppression. Then, a threshold is set to keep only proposals having high confidence scores. In this study, this threshold is set to 0.5.

We evaluate the performance of the trained residential building detector using the testing image set. Precision-recall (PR) and receiver operating characteristic (ROC) curves are used to evaluate the performance of the classifier quantitatively. To construct the PR curve, each detection first maps to its most overlapping ground-truth object samples. In this study, the threshold for considering a detection as successful is defined as an overlap of more than 50% IoU. True positive is defined as a detection with the highest-score (probability) mapped to each ground-truth sample, and all other detections are considered as false-positives. Precision is defined as the proportion of true positives to all detections. Recall is the proportion of the true positives to total number of ground-truth samples. Plotting the sequence of precision and recall values yields the PR curve shown in Fig. 8(a). The typical method to evaluate the performance of the object detector is to calculate the average precision (AP) (Girshick et al., 2014; Ren et al., 2015), which is the area under the PR curve (Everingham et al., 2010). The results show an AP of 85.47 % for the single class of residential house detection. An ROC curve, which represents the relationship between sensitivity (recall) and specificity (not precision), is also used to evaluate the performance of the detector. As mentioned previously, each detection is considered to be positive if its IoU ratio with its corresponding ground-truth annotation is higher than 0.5. By varying the threshold of detection scores, a set of true positives and false positives will be generated which is represented as the ROC curve. From Fig. 8(b), we can interpret this result to mean that the residential building detector consistently achieves impressive performance in terms of the ROC curve by obtaining true positive rates of 81.49% and 88.07% at 100 and 200 false positives, respectively.

Samples of detected houses using the optimal rectilinear images are shown in Fig. 9. Figures 9(a) and (b) shows typical successful cases in which all bounding boxes are correctly detected and tightly encompass each of house areas. Figures 9(c), (d), (e) and (f) show more challenging cases, which likely produce incorrect results. These results can be polished by adjusting the parameters used in the technique. Figures. 9(c) and (d) show two particular images which, in addition to correctly detecting houses, erroneous objects are also detected as a house. Figure. 9(c) demonstrates a case in which a boat is detected as a house with a score of 0.779. Also, in Fig. 9(d) a cargo trailer is detected as a house with a score of 0.571 which is not of interest in this study. As illustrated, all incorrect detections yield a low score, less than 0.8, which enables us to simply remove them using a higher threshold value for positive detection. Since in our application the target house is only likely to appear close to the center of the optimal rectilinear image, the target house would be detected with a high score, usually more than 0.95, unless either the image was captured far from the target house, or obstacles conceal the target house. Therefore, the problem of erroneous detections can be remedied by merely increasing the confidence threshold, here 0.8, to retain only the objects that receive a high score.

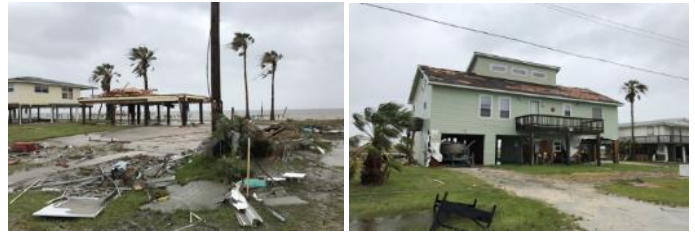

**Figure 10**. Post-disaster building reconnaissance images of two different houses after Hurricane Harvey.

Figures. 9(e) and (f) illustrate cases in which all houses are correctly detected, even though they are partially occluded by foreground objects such as trees, as with the houses in the far left and middle of the images shown in Fig. 9(e) and (f), respectively. However, the occluded detected houses are not informative or useful for inspection purposes. As is mentioned in section 4.3, the appearance of the residential house to be used for inspection purposes is a critical characteristic of the object of interest. Since our ground-truth dataset is generated based on this principle, our building detector also associates a lower score with these non-informative appearances of houses. For instance, the occluded houses shown in Fig. 9(a) and (b) are detected with scores of 0.839 and 0.507, respectively, which are considerably lower than the non-occluded houses appearing in these images. Although these objects are correctly detected as a house, the images are not useful for inspection purposes. Considering the fact that we extract several (here, 11) optimal rectilinear images for each target house, every image need not be informative for inspection, and we can safely increase the confidence threshold to remove those non-informative house detections.

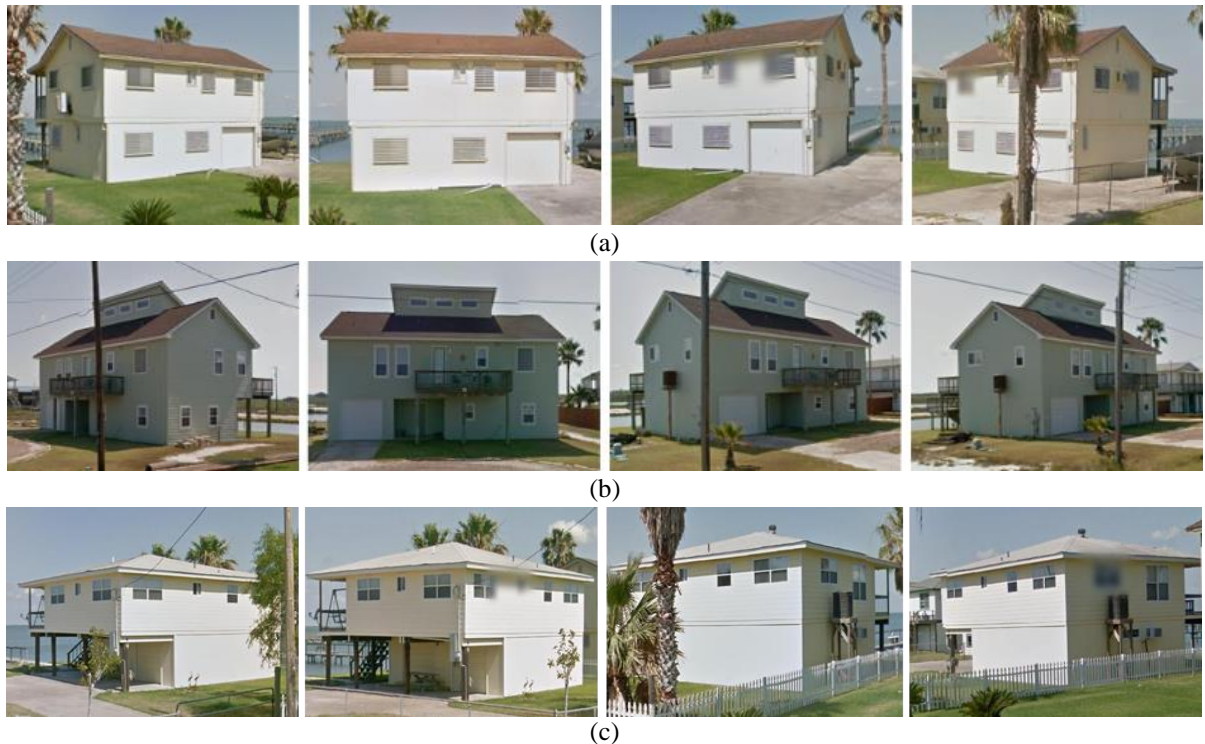

(a)

(b)

(c)

**Figure 11.** Samples of the pre-disaster house images generated from Google Street View: (a) and (b) shows the same houses in Fig. 10 (a) and (b), respectively, before Hurricane Harvey and (c), shows a house in the same neighborhood with two others but its post-disaster geo-tagged image cannot be accessed.

### 4.4 Sample implementation

In this section, the implementation of the technique developed is demonstrated using real-world post-event reconnaissance geotagged images collected from Holiday Beach, Texas after Hurricane Harvey in 2017. Pre-disaster images of two selected houses are automatically generated from the corresponding geotagged post-disaster images in Fig. 10. The house in Fig. 10(a) is merely a shell gutted by the hurricane and its appearance before the hurricane is hard to guess. The house on the right in Fig. 10(b) has significant shingle damage on the roof. One additional house is added, which does not have a geotagged image. The approximated GPS information near that second house is manually provided, rather than through a geotagged image. The locations of these houses are along the selected route in Fig. 5(a). All pre-disaster images for these three different houses are automatically generated using the technique developedThe proposed approach incorporates several program libraries including OpenCV (Bradski, 2000) and PyMap3D (Hirsch, 2018), and an implementation Faster R-CNN algorithm in TensorFlow (Chen and Gupta, 2017), and is deployed as a Python script. For this testing, we utilize the same computing resources to run this script, which are used for training the house classifier. For setting up the parameters introduced in Section 3, $n$ is set to five, so the maximum number of house images available is 11. $\theta$ and $\overline{AB}$ are 90° and 2,048 pixels, respectively, which are the same as those used for generating training images. With this setup, at each house, the approximate processing time for conducting the four main processes explained in the first paragraph of Section 3 are approximately 130, 210, 180 and 6 seconds, respectively. As mentioned in Section 4.2, high-resolution panoramas cannot be downloaded from Google Street Views directly. It takes a considerable time to create each panorama by stitching an array of the images. Unfortunately, this cannot be technically addressed in the front-end software, unless in the near future one is allowed to download panoramas from Google Street View directly.

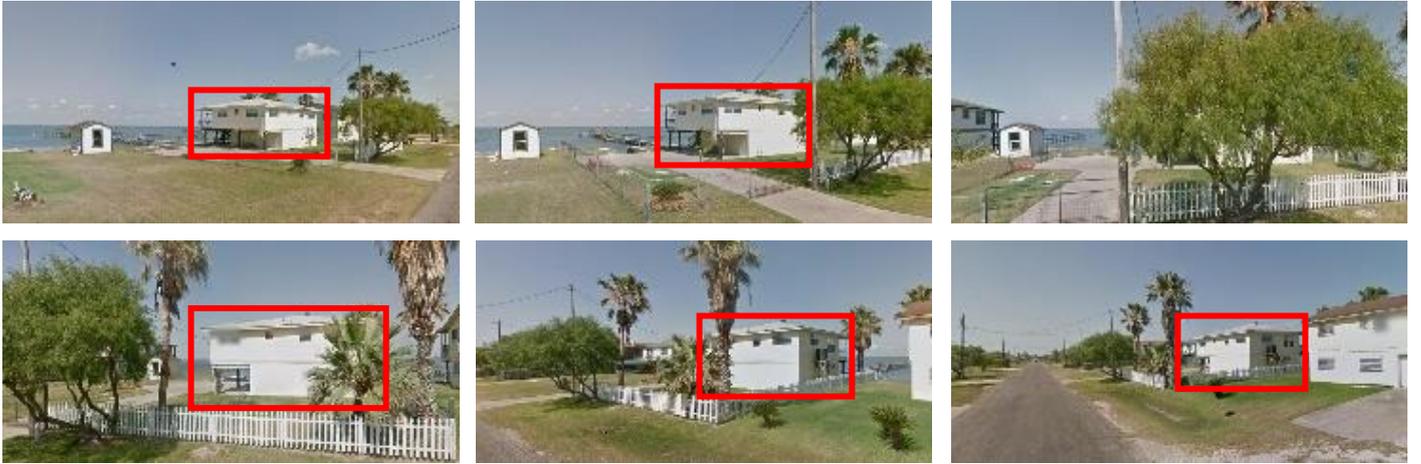

**Figure 12.** Reason to observe the house from multiple viewpoints: Optimal rectilinear images with the bounding boxes of the detected house from multiple views. Since the views of the house are partially or entirely obstructed by the foreground objects (here, trees), many images from different viewpoints should be considered to observe the entire front and side views of the house.

However, the rest of the three processes are relatively fast and can be potentially improved by exploiting better hardware or optimizing the deployment of the Python scripts.

The results in Fig. 11 show the pre-disaster house images automatically extracted from Google Street View using our technique. Figures. 11(a) and (b) show the pre-event appearance of the houses shown in Fig. 10(a) and (b), respectively, and Fig. 11(c) demonstrates the generation of pre-event images of the same house without a post-event geo-tagged image. The seven images of houses shown in Fig. 11(a) and (b) and the five images of houses shown in Fig. 11(c), from various viewpoints are detected from the 11 optimal rectilinear images, although only four representative samples of those images are provided in Fig. 11. It is clear from these samples that including images of each house from multiple viewpoints is important for enabling observation of the entire side and front façade of the house. Clearly, they provide valuable information about the pre-hurricane state of the house. For example, the pre-event appearance of the house shown in Fig.11(a) can hardly be imagined using only the post-disaster image in Fig. 10(a).

Figure 12 shows the six optimal rectilinear images generated for the house in Fig. 11(c). The bounding box of the corresponding house in each rectilinear image is marked. This example further demonstrates why multiple images from different viewpoints should be extracted (in other words, $n$ should be set to more than two). Since large trees in front of this house block its view, multiple images are needed to observe the entire view of the house. This situation commonly occurs in cluttered scenes where various sources of foreground objects are potentially present in a street view, for instance, large street signs, parked vehicles, trees, or pedestrians. This issue can be minimized by increasing $n$ to consider more viewpoints.

### 4.5 Technique validation

For further evaluation of the technique developed, we introduce a new metric, denoted *Overall Practicality (OP)*. In this technique, the potential sources of error include: (1) incorrect estimation of the GPS location of the building due to a feature matching failure (Steps 3~ 8), (2) false building detection on the rectilinear images using the trained classifier, and (3) incorrect GPS records for the street view panoramas. The new metric, $OP$, is designed to quantitatively evaluate the likelihood of extracting a sufficient set of building images of the quality of the resulting images in Fig. 11. $OP$ is formulated as:

$$OP = \frac{N_u}{N_t - N_o} ,$$
(4)

where, $N_u$ is the number of extracted images that contain a satisfactory view of the target building on the corresponding rectilinear images. When the extracted building is not placed near the center of the rectilinear images, due to the projection direction error, it does not contribute to $N_u$ although it contains enough of the building's appearance. Also, if the trained classifier only detects a portion of the building or does not tightly estimate its region with a bounding box, it is also not included in $N_u$. $N_t$ is the number of all building images which can be extracted from the available panorama. It can be calculated as $N_t = N_b(2n + 1) - \sum_{i=1}^{N_b} m_i$, where $N_b$ is the total number of target buildings, $n$ is the maximum allowable number of images (see Section. 5.2), and $m_i$ is the number of missing panoramas at building $i$. As mentioned in Section 3 (Step 2), if a building is located close to a dead-end or a cul-de-sac, street view panoramas are not available. $N_o$ is the number of occluded images. The occlusion due to foreground objects (e.g. tree, car, or fence) is inevitable, and can obstruct the view of the building. Thus, it is a clear basis for needing multi-view images. This number is not contained in either $N_u$ or $N_t$.

To evaluate the performance of the technique developed using $OP$, we randomly select 50 residential buildings along a different street (marked using a black line in Fig. 5(a)). In the actual implementation, the input of the technique is a geo-tag image which must be captured close to the building-of-interest. However, existing data sets rarely have a sufficient number of geo-tagged images available. Thus, in this evaluation, we manually provide similar GPS information for each building with the assumption that these GPS data can be obtained from geo-tagged post-disaster images during actual usage of the technique (step 1 in Fig. 3). Recall that this GPS information corresponds to a location on the street close to each building, and not the location of the buildings. Having this GPS information, we input GPS information to Step 2 and execute the rest of the steps. The performance is quite successful, and we obtain an $OP$ of 0.838. Among 50 buildings, only one building has three missing panoramas, and thus, $N_t = 547$ and $N_o$ and $N_u$ are 39 and 426, respectively.

## 5 Summary and Conclusions

This study presents an automated technique to extract pre-event building images from street view panoramas, publicly available online. Although a great many images are being collected from damaged buildings after disasters to learn important lessons from those events, only limited knowledge can be obtained by using such images without providing suitable information regarding its pre-disaster state. A visual comparison of the building before and after a disaster is crucial to trace the cause of a failure or the reason for damage during disasters. The technique developed herein automates the process of extracting building images from existing street view panoramas to support the building reconnaissance process. Once a user provides an approximate location of a building through a geotagged image, several high-quality external views of the entire building are rapidly generated for use.

The performance of the technique developed here is successfully demonstrated using actual reconnaissance images as well as Google Street View panoramas collected from Holiday Beach and Rockport, Texas, which suffered significant damage during Hurricane Harvey in 2017. A total of 100 houses, 1065 images at Holiday Beach captured with the Google Street View panoramas are manually annotated to build a robust house classifier using a region-based convolutional neural network algorithm. The performance of the house classifier is evaluated using an independent set of test images, yielding average precision values of 85.47%. An actual implementation of the technique is performed and discussed using three post-disaster damaged houses collected at Holiday Beach in Rockport. These images were collected by field investigators during a post-disaster reconnaissance mission. All of the pre-disaster images, corresponding to the house in each post-disaster image, are successfully extracted from Google Street View panoramas. The technique represents a promising example of how to easily and automatically add value to both newly collected and existing (legacy) volumes of visual data.

## ACKNOWLEDGMENTS

The authors wish to acknowledge support from Purdue Center for Resilient Infrastructures, Systems, and Processes (CRISP) and National Science Foundation under Grant No. NSF 1608762. We would like to give special thanks to Thomas Smith at TLSmith Consulting Inc., and Tim Marshall at HAGG Engineering Co. for their advice and image contributions.